\documentclass{article} % For LaTeX2e
\usepackage{iclr2023_conference,times}

\usepackage{amsmath,amsfonts,bm}

\def\figref#1{figure~\ref{#1}}
\def\Figref#1{Figure~\ref{#1}}

\def\secref#1{section~\ref{#1}}
\def\Secref#1{Section~\ref{#1}}
\def\eqref#1{equation~\ref{#1}}
\def\Eqref#1{Equation~\ref{#1}}
\def\1{\bm{1}}

\DeclareMathAlphabet{\mathsfit}{\encodingdefault}{\sfdefault}{m}{sl}
\SetMathAlphabet{\mathsfit}{bold}{\encodingdefault}{\sfdefault}{bx}{n}

\usepackage{hyperref}
\usepackage{url}

\usepackage{subcaption}
\usepackage[]{caption}	% makes captions ragged right - thanks to Bryce Lobdell
\usepackage{lscape}                                         % Useful for wide tables or figures.
\usepackage[export]{adjustbox}

\usepackage[lined,ruled,linesnumbered]{algorithm2e}

\usepackage{booktabs}                   % Publication quality tables
\usepackage{multirow}

\usepackage{paralist}
\usepackage{enumitem}

\usepackage{bm}                          % Make bold, italic math symbols
\usepackage{epsfig}                      % for figures
\usepackage{times}
\usepackage{pifont}

\usepackage{units}
\usepackage{url}  % Hyphenation of URLs.
\usepackage{xspace}
\usepackage[percent]{overpic}
\usepackage{ifthen}
\newcommand{\networkName}{GPR-Net} % name of network

\def\etal{et al.~}			  % and others, and co-workers
\def\eg{e.g.,~}               % for example
\def\ie{i.e.,~}               % that is, in other words
\newlength\paramargin
\newlength\figmargin
\newlength\secmargin
\newlength\figcapmargin

\newcommand{\heading}[1]
{
\vspace{1mm}
\noindent \textbf{#1}
}   

\newcommand{\tabref}[1]{Table~\ref{tab:#1}}

\long\def\ignorethis#1{}
\newcommand {\james}[1]{{\color{violet}\textbf{James: }#1}\normalfont}
\newcommand {\eric}[1]{{\color{magenta}\textbf{Eric: }#1}\normalfont}

\newcommand {\hl}[1]{{\color{red}#1}}

\newcommand{\tb}[1]{\textbf{#1}}

\def\panofirst{{\mathbf{I^1}}}
\def\panosecond{{\mathbf{I^2}}}
\def\panouinput{u_i}
\def\panouinputset{{\mathbb{U}}}

\def\coroutput{o_i}
\def\coroutputset{{\mathbb{O}}}
\def\coroutputgt{\overline{o}_i}
\def\coroutputgtset{\mathbb{\overline{O}}}

\def\covisoutput{c_i}
\def\covisoutputset{{\mathbb{C}}}
\def\covisoutputgt{\overline{c}_i}
\def\covisoutputgtset{\mathbb{\overline{C}}}
\def\layoutfirstceiling{v_i^1}
\def\layoutfirstceilingset{{\mathbb{V}^1}}
\def\layoutfirstfloor{\hat{v}_i^1}
\def\layoutfirstfloorset{{\mathbb{\hat{V}}}^1}
\def\layoutsecondceiling{v_i^2}
\def\layoutsecondceilingset{{\mathbb{V}^2}}
\def\layoutsecondfloor{\hat{v}_i^2}
\def\layoutsecondfloorset{{\mathbb{\hat{V}}}^2}
\def\horizondepthfirstceiling{{\mathbb{D}^1}}
\def\horizondepthsecondceiling{{\mathbb{D}^2}}
\def\horizondepthfirstfloor{{\mathbb{\hat{D}}^1}}
\def\horizondepthsecondfloor{{\mathbb{\hat{D}}^2}}
\def\horizondepthfirstgt{\overline{\mathbb{D}^1}}
\def\horizondepthsecondgt{\overline{\mathbb{D}^2}}
\def\layoutfirstpointset{{\mathbb{P}^1}}
\def\layoutfirstpointsetmatch{{\mathbb{\hat{P}}^1}}
\def\layoutsecondpointset{{\mathbb{P}^2}}
\def\layoutsecondpointsetwarp{{\mathbb{P}^2_{warrped}}}
\def\layoutfirstpoint{p_i^1}
\def\layoutfirstpointmatch{\hat{p}_i^1}
\def\layoutsecondpoint{p_i^2}

\def\translation{T}
\def\rotation{R}

\def\layoutset#1#2{{\mathbb{V}^{#2}_{#1}}}

\def\horizondepthset#1#2{{\mathbb{D}^{#2}_{#1}}}
\def\horizondepthsetgt#1#2{{\mathbb{\overline{D}}^{#2}_{#1}}}

\def\mlplayoutceiling#1{{\mathcal{F}^{c}_{#1{}}}}
\def\mlplayoutfloor#1{{\mathcal{F}^{f}_{#1{}}}}

\def\angularErr#1{\Delta#1}

\def\mlpfirstceilingparam{\Phi_{\layoutfirstceilingset}}
\def\mlpfirstceiling{{\mathcal{F}_{\mlpfirstceilingparam}}}
\def\mlpfirstfloorparam{\Phi_{\layoutfirstfloorset}}
\def\mlpfirstfloor{{\mathcal{F}_{\mlpfirstfloorparam}}}
\def\mlpsecondceilingparam{\Phi_{\layoutsecondceilingset}}
\def\mlpsecondceiling{{\mathcal{F}_{\mlpsecondceilingparam}}}
\def\mlpsecondfloorparam{\Phi_{\layoutsecondfloorset}}
\def\mlpsecondfloor{{\mathcal{F}_{\mlpsecondfloorparam}}}
\def\mlpcoroutputset{{\mathcal{F}^{cor}}}
\def\mlpcovisoutputset{{\mathcal{F}^{covis}}}
\def\transformer{{\mathcal{F}^{T}}}

\def\losslayout{\mathcal{L}_{layout}}
\def\losscorrespondence{\mathcal{L}_{cor}}
\def\losscovisibility{\mathcal{L}_{covis}}
\def\losscyclecovis{\mathcal{L}^{covis}_{cycle}}
\def\losscyclecorr{\mathcal{L}^{cor}_{cycle}}
\def\losscycleconsistency{\mathcal{L}_{cycle}}
\def\losstotal{\mathcal{L}_{total}}

\definecolor{gtlayout}{RGB}{47, 82, 224}
\definecolor{ourlayout}{RGB}{188, 237, 9}
\definecolor{ledlayout}{RGB}{249, 203, 64}
\definecolor{lgtlayout}{RGB}{255, 113, 91}

\title{GPR-Net: Multi-view Layout Estimation via a Geometry-aware Panorama Registration Network}
\author{
Jheng-Wei Su$^1$, Chi-Han Peng$^2$, Peter Wonka$^3$, Hung-Kuo Chu$^{4}$\\
$^{1,4}$ National Tsing Hua University\\
$^2$ National Yang Ming Chiao Tung University\\
$^3$ King Abdullah University of Science and Technology (KAUST) \\
\texttt{$^1$jhengweisu@gapp.nthu.edu.tw, $^2$pengchihan@nycu.edu.tw}\\
\texttt{$^3$pwonka@gmail.com, $^4$hkchu@cs.nthu.edu.tw}
}

\iclrfinalcopy % Uncomment for camera-ready version, but NOT for submission.
\begin{document}

\maketitle

\begin{abstract}
%====
% What kind of problem do we want to solve, and why is it a challenging problem?
%====
%Reconstructing 3D layouts from multiple $360^{\circ}$ panoramas has received increasing attention recently as it paves the way toward estimating a complete layout of a large-scale and complex room.
Reconstructing 3D layouts from multiple $360^{\circ}$ panoramas has received increasing attention recently as estimating a complete layout of a large-scale and complex room from a single panorama is very difficult.
%
%However, jointly estimating the layout geometry and registration among panoramas is non-trivial. The problem is even harder when two panoramas are far away from each other.
%
%====
% How do previous methods handle the problem? What are their limitations?
%====
The state-of-the-art method, called PSMNet~\citep{Wang_2022_CVPR}, introduces the first learning-based framework that jointly estimates the room layout and registration given a pair of panoramas.
However, PSMNet relies on an approximate (\ie "noisy") registration as input. Obtaining this input requires a solution for \emph{wide baseline registration} which is a challenging problem.
%
%To tackle this \emph{wide baseline registration} problem, a state-of-the-art method~\cite{Wang_2022_CVPR} assumed an approximate (i.e., "noisy") relative pose between pair of panoramas is given and learns from datasets to refine the initial pose.  
%
%Such an assumption may be impractical in real-world scenarios as reliably estimating approximate poses from two panoramas remains a challenging problem in the wide baseline setting. 
%
%Both conventional Structure-from-Motion (SfM) or learning-based regression methods tend to fail in the wide baseline setting.
%
%====
% What is our key idea?
%====
In this work, we present a complete multi-view panoramic layout estimation framework that jointly learns panorama registration and layout estimation given a pair of panoramas without relying on a pose prior.
%deep learning framework for estimating a 3D layout from a pair of panoramas without relying on a pose prior.
%In this work, we present a novel and deep learning framework for estimating a 3D layout from a pair of panoramas without relying on a pose prior.
%
The major improvement over PSMNet comes from a novel Geometry-aware Panorama Registration Network or \networkName ~that effectively tackles the wide baseline registration problem by exploiting the layout geometry and computing fine-grained correspondences on the layout boundaries, instead of the global pixel-space.
%Our key insight is that by exploiting the geometric features of layouts (\ie corners and boundaries), one can reliably estimate the panorama registration in \hl{3D} via establishing the corner-to-corner and wall-to-wall correspondence between two layouts.
%
%Therefore, we devise an end-to-end architecture that jointly learns layout registration and layout prediction.
%
Our architecture consists of two parts. First, given two panoramas, we adopt a vision transformer to learn a set of 1D horizon features sampled on the panorama. These 1D horizon features encode the depths of individual layout boundary samples and the correspondence and covisibility maps between layout boundaries. We then exploit a non-linear registration module to convert these 1D horizon features into a set of corresponding 2D boundary points on the layout. Finally, we estimate the final relative camera pose via RANSAC and obtain the complete layout simply by taking the union of registered layouts.
%Given two input panoramas, our model adopts a vision transformer to learn depth, correspondence, and co-visibility features along the layout boundaries. \hl{All these features are represented as 1D horizon lines so the model can benefit....}\james{What are the advantages of using horizon line representation?} 
%
Experimental results indicate that our method achieves state-of-the-art performance in both panorama registration and layout estimation on a large-scale indoor panorama dataset ZInD~\citep{Cruz_2021_CVPR}.
\end{abstract}

%This representation has the advantages of having more elements to register (\eg 256 samples per panorama), more supervision signal for fine-grained estimation, and thus leading to better learning performance.

\if 1
%We exploit the vision transformer architecture to encode low-level geometric features as a set of horizon lines representing single-view layout depth and correspondence between two panoramas. The relative pose between two panoramas is then derived from depth and correspondence feature lines, and is used to fuse two partial layouts into a complete one.  
%Instead of computing full-image pixel correspondence between panoramas like conventional Structure-from-Motion (SfM), or directly regress pose parameters using features learned from panoramas, our key idea is to leverage the geometric features of layout (\ie, corners, wall-ceiling and wall-floor boundaries) to estimate the correspondence between two panoramas. \james{why does this idea work?}
%
%====
% What is the goal of this project?
%====
In this paper, we present a novel deep learning framework for estimating the 3D layout of an indoor scene from a pair of $360^{\circ}$ panoramas.
%
%====
% What does our architecture do? What components make our architecture special
%====
Our system is an end-to-end architecture that jointly learns registration and layout prediction from input equirectangular views. We exploit the vision transformer architecture to encode low-level geometric features as a set of horizon lines representing single-view layout depth and correspondence between two panoramas. The relative pose between two panoramas is then derived from depth and correspondence feature lines, and is used to fuse two partial layouts into a complete one.  
%
%====
% Why our architecture is better than previous works?
%====
Our architecture has several advantages compared to existing solutions. 
% Our architecture has several advantages compared to previous state-of-the-art method (PSMNet).
%Jointly estimating low-level depth and correspondence features has several advantages compared to existing approaches that directly regress the pose parameters.
%
First, in contrast to a perspective ceiling view, an equirectangular panorama contains richer content for feature extraction.
Second, instead of learning registration and layout estimation using separate networks, our horizon- depth and correspondence representation enables a unified architecture for joint learning.
Therefore, our model is capable of robustly predicting layout geometry and pose parameters without using a (noisy) pose prior.
As a result, our method achieve state-of-the-art performance in both panorama registration and 3D layout estimation on a large-scale indoor panorama dataset.

%Existing approaches directly regress the pose parameters and thus require approximate registration as input to regularize the unknown scale issue in conventional stereo matching 
%
%Existing approaches require approximate registration as input and 
%during both train, either manually specified or estimated from stereo matching method, as input, which largely restricts the scope of application in practice. 
%
%Our quantitative results show an improvement over the current state-of-the-art registration algorithm and an improvement over single-view room layout estimation algorithms.

Input relative pose may be only approximately known (represented by the noisy alignment)
PSMNet joinly estimate the complete visible room layout in 2D, while refining a given noisy relative pose
Wide baseline 2-view Structure from Motion (SfM) is still an open problem. PSMNet assumes that an input pose, potentially noisy, is provided. For example, this could be based on a rough user input [4] or matching corresponding semantic elements with noisy predictions [29].
PSMNet is a joint pose-layout deep architecture to predict 2D room layout and refine a noisy \hl{3 DOF relative camera pose} in an end-to-end manner.
PSMNet is a 2-view layout reconstruction with approximate poses.

\james{Texts from ICCV submission}
We propose a novel framework to estimate room layouts from multiple panoramic RGB images (i.e., panoramas) taken inside the same room.
Our solution consists of the following major components. First, we propose a new registration method for estimating a rigid transformation between the camera locations of multiple panoramas. Second, we propose a joint segmentation architecture that can jointly segment multiple registered panoramas. Third, we propose a graph-cut based binary segmentation that produces room layouts with sharp corners and straight walls. To enable this research we also introduce a consistently annotated multi-view room layout dataset. Our quantitative results show an improvement over the current state-of-the-art registration algorithm and an improvement over single-view room layout estimation algorithms.

% Eric's abstract
We present a vision transformer framework for indoor stereo panorama layout reconstruction. Although the previous methods try to reconstruct the layout using the pose prior, the predicted poses from other existing methods are still unreliable. In detail, the global scale of estimated poses derived from traditional methods is unknown, which usually causes the layout fusion to be unstable. To achieve better stereo panorama layout reconstruction, we propose a unified network to predict both the layout prediction and pose registration jointly. Specifically, we take the estimated single view layout prediction as a prior to guide the registration process for predicting the relative pose containing the global scale of single view layout. We then fusion the single view layout with the predicted poses. Experimental results show that our work outperforms the current state-of-the-art methods on a public panoramic dataset.
\fi
\section{Introduction}
\label{sec:intro}

% Some of the problems come from PSMNet
% They need initial noisy pose
% Why do they need initial noisy pose? Their SP^2 network may need enough co-visible area to do the registration. What if the initial pose is so bad? The low-overlapping cases decrease more than 10% on 2D IoU between w/ and w/o GT pose. 
% 1. They project two panorama into the ceiling view
% 2. They do some registration on the ceiling view
% 3. They use those refined pose on the remaining process
% What are our adv.?
% 1. We use the equirectangular view as input which could largely increase our information
% 2. We predict co-visibility mask as well to increase our registration accuracy and make sure we only consider the correspondence on the co-visible area
% 3. Leveraging the beneficial of predicting layout using depth representation, we extend the horizon-depth with predicting correspondence and co-visibility. (LED2-net has shown that this kind of layout representation is much better than the segmentation based method.)

% ================================================================
% \paragraph{What's the problem?}
% % Describe the input and output.
% \paragraph{Why we should care about the problem?}
% % A step further to an existing problem? Applications?
% ================================================================

In this paper, we tackle the problem of room layout estimation from multiple $360^{\circ}$ panoramas. 
Many approaches that can estimate room layouts from a single panorama have been proposed~\citep{zou2018layoutnet,Yang_2019_CVPR,sun2019horizonnet,pintore2020atlantanet}.
However, these methods did not take advantage of "multi-view" data in which multiple panoramas are taken to better capture a single room.
These kinds of data are actually common as evidenced by several indoor datasets such as ZInD~\citep{Cruz_2021_CVPR}, Matterport3D~\citep{chang2017matterport3D}, Gibson~\citep{xiazamirhe2018gibsonenv}, and Structure3D~\citep{Structured3D} in which photographers often take multiple panoramas to better capture complex, non-convex rooms that would be partially occluded from just a single location.
%
% \james{@Eric, please check the name of dataset and add citation, }

% ================================================================
% \paragraph{What other people have done related to this problem?}
% % What are the main challenges?
% \paragraph{Why existing approaches are not satisfactory?}
% % If they are satisfactory, you don't need to do this research.
% ================================================================

Our work mainly improves upon a recent paper, PSMNet~\citep{Wang_2022_CVPR}, that tackled the problem of layout estimation from two panoramas captured in the same room.
The idea of PSMNet is to build an architecture that first registers two panoramas in their ceiling view projections and then jointly estimates a 2D layout segmentation.
An important aspect of their architecture is that the layout estimation and registration can be trained jointly.
However, a major limitation of PSMNet (also mentioned in their paper) is that the architecture relies on an initial approximate registration. The authors argued that such an approximate registration could be given either manually or computed by external methods such as Structure from Motion (SfM) methods or~\citet{Shabani_2021_ICCV}.
While a manual registration may work, the method would no longer be automatic.
When experimenting with existing methods for approximate registration, we observed that they frequently make registration errors and even fail to provide a registration in a substantial number of cases.
The main reason is that the required registration mainly falls into the category of wide baseline registration with only two given images.
For example, our results show that the state-of-the-art SfM method OpenMVG~\citep{moulon2016openmvg} fails to register $76\%$ of panorama pairs from our test dataset.
It is thus impractical to assume an independent algorithm that can reliably give an approximate solution to the challenging wide baseline registration problem.
In addition, relying on such an algorithm moves a critical part of the problem to a pre-process.

% ================================================================
% \paragraph{What our idea is? Why it is great?}
% % Start with ``In this paper, we propose to ...'' to show a clear distinction between previous work and this paper.
% ================================================================

\begin{figure*}[!t]
    % \begin{subfigure}[!t]{\linewidth}
    %   \includegraphics[width=\linewidth,keepaspectratio]{figure/images/overview.jpg}
    % \end{subfigure}
    \begin{overpic}[width=\textwidth]{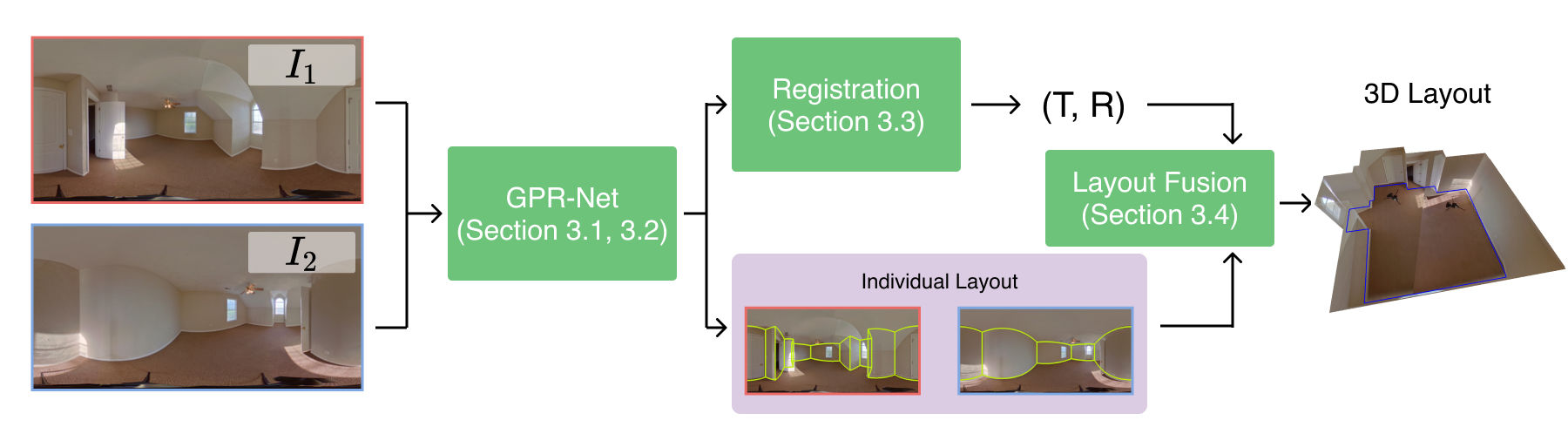}
    
    %\put(38.5, 24.5){\begin{minipage}{0.25\textwidth}\centering Conditional \\ Image \\ Generator  ${G_\theta}$ \end{minipage}}
    %\put(39, 7){\begin{minipage}{0.25\textwidth}\centering  Appearance \\ Field  ${F_\theta}$ \end{minipage}}
    %\put(55, 15.5){\begin{minipage}{0.5\textwidth}\centering ${L_{1}} + {L_{perceptual}} + {L_{gradient}}$\end{minipage}}
    % + Gradient smooth loss \\  (\secref{loss_function})\end{minipage}}
    %\put(3.5, 19){\begin{math} (D_1, S_1) \end{math}}
    %\put(15.5, 19){\begin{math} (D_2, S_2) \end{math}}
    %\put(24.5, 19){....}
    %\put(28.8, 19){\begin{math} (D_v, S_v) \end{math}}
    %\put(64.3, 20.5){\begin{math} G(D_1, S_1) \end{math}}
    %\put(76.5, 20.5){\begin{math} G(D_2, S_2) \end{math}}
    %\put(70, 18.7){Photorealistic panoramic images}
    %\put(86.7, 20.5){...}
    %\put(90.5, 20.5){\begin{math} G(D_v, S_v) \end{math}}
    %\put(31.2, 11.5){ $\appearanceembed$ \begin{math}\in \mathbb{R}^8 \end{math}}
    %\put(31, 7.3){\begin{math} s \in [1, k] \end{math}}
    %\put(30.5, 3){\begin{math} r(X, Y, Z) \end{math}}
    % \put(9.5, 0){\begin{minipage}{0.5\textwidth}\centering (Section \ref{sec:field_representation}, \ref{sec:appearance_embedding}) \end{minipage}}
    %\put(4.7, -1.8){\begin{minipage}{0.25\textwidth}\centering Input Scene \end{minipage}}
    %\put(68, 1){\begin{minipage}{0.25\textwidth}\centering Synthesized results \end{minipage}}
    \end{overpic}
    \\
    \caption{\tb{Proposed multi-view panorama registration and layout estimation framework.} Given two panorama images, a neural network (GPR-Net) jointly predicts layout boundary correspondences and individual layouts for the two panoramas (\secref{sec:network_archi},\secref{sec:loss_functions}). The correspondences are fed to a registration module to compute the relative camera pose $(T,R)$ (\secref{sec:registration}). A layout fusion module then computes a unified 3D layout given the camera pose and the individual layouts (\secref{sec:fusion}).
    }
\label{fig:pipeline}
\end{figure*}

%We therefore set out to make an improvement to the registration part of the multi-view layout estimation problem.
%
Therefore, we set out to develop a complete multi-view panorama registration and layout estimation framework that no longer relies on an approximate registration given as input as shown in~\Figref{fig:pipeline}.
%
%To achieve this, we propose a novel deep learning architecture based on the following design ideas:
To achieve this, we propose a novel Geometry-aware Panorama Registration Network, or \networkName,~based on the following design ideas.
First, our experiments indicate that a global (pixel-space) registration that directly regresses pose parameters (\ie translation and rotation) is too ambitious.
Instead, we propose to compute more fine-grained correspondences in a different space.
Specifically, \networkName~conceptually samples the layout boundaries of two input layouts and computes features for the sampled locations. For each boundary sample in each of the two panoramas, it estimates the distance from the camera (depth). In addition, it estimates the correspondence map from the samples in the first panorama to the second panorama and a covisibility map describing if a sample in the first panorama is visible in the second panorama.
%
%These feature maps (depth, correspondence, and covisibility) are encoded as a set of compact 1D horizon lines sampled on the panorama.
Each of these maps (depth, correspondence, and covisibility) is a 1D sequence of values.

%
%The correspondence between two layouts is represented as a compact 1D horizon line sampled on the panorama.
%
This representation has the advantages of having more elements to register (\eg 256 samples per panorama) and more supervision signal for fine-grained estimation. This leads to better learning performance.
Second, we build a non-linear registration module to compute the final relative camera pose.
%that combines the estimated horizon-depth, horizon-correspondence, and horizon-co-visibility maps to compute the final pose parameters.
%
The module combines two horizon-depth maps with the horizon-correspondence and horizon-covisibility maps to obtain a set of covisible corresponding boundary samples in a 2D coordinate system aligned with the ceiling plane, followed by a RANSAC-based pose estimation.
Note that this non-linear space is more expressive and can encode a richer range of maps between two panoramas.
The final complete layout is obtained simply by taking the union of two registered layouts.

%
%Our idea is to move the correspondence problem to the space of boundary segments and to compute correspondences between individual boundary segments. This has the advantage that we have more elements to register (e.g. 256 boundary segments in each image), more supervision signal for fine grained supervision, and thereby better learning.
%
%Second, we build a two-stage registration framework. Instead of directly computing a linear registration, we first compute a non-linear registration of boundary elements. This non-linear space is more expressive and can encode a richer range of maps between two panoramas. Only in the second stage, we aggregate the non-linear correspondences into a linear map using L2D (XXX please fix).

% ================================================================
% \paragraph{How well are the proposed method? What kinds of experiments have you done for validation?}
% % Start with ``To validate our method...'' to summarize what kinds of experiments you have done to prove the work is really good.
% ================================================================
We extensively validate our model by comparing with the state-of-the-art panorama registration method and multi-view layout estimation method on a large-scale indoor panorama dataset ZInD~\citep{Cruz_2021_CVPR}.
The experimental results demonstrate that our model is superior to competing methods by achieving a significant performance boost in both panorama registration accuracy (mAA$@5^{\circ}$: $+68.5\%$(rotation), $+63.0\%$(translation), mAA$@10^{\circ}$: $+74.1\%$(rotation), $+72.3\%$(translation)) and layout reconstruction accuracy (2D IoU $+4.5\%$). 
%

% ================================================================
% \paragraph{List of contributions}
% % Create a list of contributions. This helps the reviewers to summarize the paper. The rest of the paper provide evidences for the claimed novelty/contributions. Use forward references to provide a roadmap to the paper, e.g., we propose XXX (\secref{algorithm})
% ================================================================
% \input{iclr2023/Figures/fig_pipeline}
In summary, our contributions are as follows:
\begin{itemize}
    \item We propose the first complete multi-view panoramic layout estimation framework. Our architecture jointly learns the layout and registration from data, is end-to-end trainable, and most importantly, does not rely on a pose prior.
    %We propose the first complete joint layout and correspondence prediction for multi-view panoramas. In contrast to previous work PSMNet, our framework does not rely on a pose prior. % compare with PSMNet
    \item We devise a novel panorama registration framework to effectively tackle the wide baseline registration problem by exploiting the layout geometry and computing a fine-grained correspondence of samples on the layout boundaries.
    %we provide a new SOTA wide baseline registration algorithm for panorama images
    \item We achieve state-of-the-art performance on ZInD~\citep{Cruz_2021_CVPR} dataset for both the stereo panorama registration and layout reconstruction tasks.
    % talk about we achieve the best result on ZInD
\end{itemize}
\section{Related Work}
\label{sec:related}

%Extreme Structure from Motion for Indoor Panoramas without Visual Overlaps

\subsection{Single-view room layout estimation}
%\paragraph{Single-view room layout prediction.} 
There exist many methods to estimate the room layouts from just a single image taken inside an indoor environment. Methods that take only one perspective image include earlier attempts that relied on image clues and optimization~\citep{5459411,Hoiem2007,Ramalingam2013Lifting3M} and later neural networks~\citep{roomnet,8962030}. Capturing the increasing availability and popularity of full $360^{\circ}$ panoramic images, the seminal work by Zhang~\etal~\citep{panocontext} proposed to take panoramas as native inputs for scene understanding. Recently, several methods were proposed to predict the room layouts from a single panorama using neural networks. A major difference between these methods is the assumption on the shape of the room layouts - from being strictly a cuboid~\citep{zou2018layoutnet}, Manhattan world~\citep{Yang_2019_CVPR,sun2019horizonnet}, to more recently general 2D layouts (Atlanta world)~\citep{pintore2020atlantanet}. For our work, we choose to adopt the Manhattan assumption because more corresponding data is available. See~\citet{zou2021mp3dlayout} for a thorough survey on predicting Manhattan room layouts from a single panorama. More recent methods delivered state-of-the-art performance by transforming the problem into a depth-estimation one~\citep{Wang_2022_CVPR} or by leveraging powerful transformer-based network architecture~\citep{Jiang_2022_CVPR}.
Although these single-view methods perform well in the cuboid and L-shape rooms, they tend to fail in the large-scale, complex and non-convex rooms where a single-view panorama covers only part of the whole space due to occlusion. 

\subsection{Panorama registration}
%\paragraph{Panorama registration.} 
Image registration, i.e., finding transformations between the cameras of two or multiple images taken of the same scene, is a key component of Structure-from-Motion (SfM). See~\citet{SurveySfM} for a recent survey and~\citet{10.5555/861369} for an extensive study. Registration problems can be categorized by: 1) the assumptions about the camera model, e.g., perspective (pinhole camera), weak-perspective, or orthographic, 2) the assumptions about the transformation, e.g., rigid, affine, or general non-rigid, and 3) the types of the image inputs, e.g., perspective images or full $360^{\circ}$ panoramas, and with/without depths. In addition, the difficulty differs greatly on whether the images are taken densely or sparsely. Modern takes on registration problems often leverage state-of-art programs/libraries such as COLMAP~\citep{schoenberger2016colmap} and OpenMVG~\citep{moulon2016openmvg}. Our problem falls into a lesser-studied category: {\em registering rigid transforms between sparse panoramas}. While there exist methods that tackle sparse perspective image inputs~\citep{8374572,Fabbri_2020_CVPR} and methods that handle panoramas natively~\citep{6130266,10.1109/3DIMPVT.2012.45,JI2020169}, our results show that we can improve upon the state-of-the-art panorama registration methods in our sparse view setting. A key bottleneck is that traditional SfM methods fail to handle the wide baseline registration problem where the views are far apart from each other. In~\citet{Shabani_2021_ICCV}, SfM of extremely sparse panoramas was tackled by matching room types and specific elements such as doors and windows. 
%In our work, we propose a new registration approach that can handle such challenging cases without the needs to identify room types or specific elements.
In contrast to previous methods that perform the registration in the global pixel-space, we propose a novel learning-based panorama registration framework that directly compute the registration between two panoramas without taking any prior knowledge as input.
Our method may have some similarities to the ECCV 2022 paper~\citep{covispose}, but the paper was not available online when we developed our method or wrote the first drafts of our paper.
%exploits the layout geometry to compute fine-grained correspondences on the layout boundaries.
%Our model therefore can directly compute the registration between two panoramas without taking any prior knowledge as input.

\subsection{Scene reconstruction using sparse panoramas}
%\paragraph{Scene reconstruction using sparse panoramas.}
Attempts to reconstruct indoor scenes using just a handful of RGB panoramas as inputs \cite{Pintore20183DFP,https://doi.org/10.1111/cgf.13842} are nascent but promising since photographers are adapting $360^{\circ}$ cameras into their workflows (e.g., Matterport 3D capture system~\citep{chang2017matterport3D}) and it is awkward to capture dense panoramic inputs due to camera/tripod setups. 
While previous methods assume that all the input panoramas are already registered, the PSMNet~\citep{Wang_2022_CVPR} introduces the first learning-based framework that jointly estimates the room layout and registration given a pair of panoramas. However, it still has a major bottleneck that an initial approximate (noisy) registration must be given (\eg either manually specified or computed by external methods) during both the training and inference stages.
Our {\networkName} is also an end-to-end deep neural network that jointly learns the room layout and panorama registration. Most importantly, our model does not rely on a pose prior and is thus suitable for real-world application scenarios.
% \clearpage
\section{Methodology}
\label{sec:method}
\begin{figure*}[!t]
    % \begin{subfigure}[!t]{\linewidth}
    %   \includegraphics[width=\linewidth,keepaspectratio]{figure/images/overview.jpg}
    % \end{subfigure}
    \begin{overpic}[width=\textwidth]{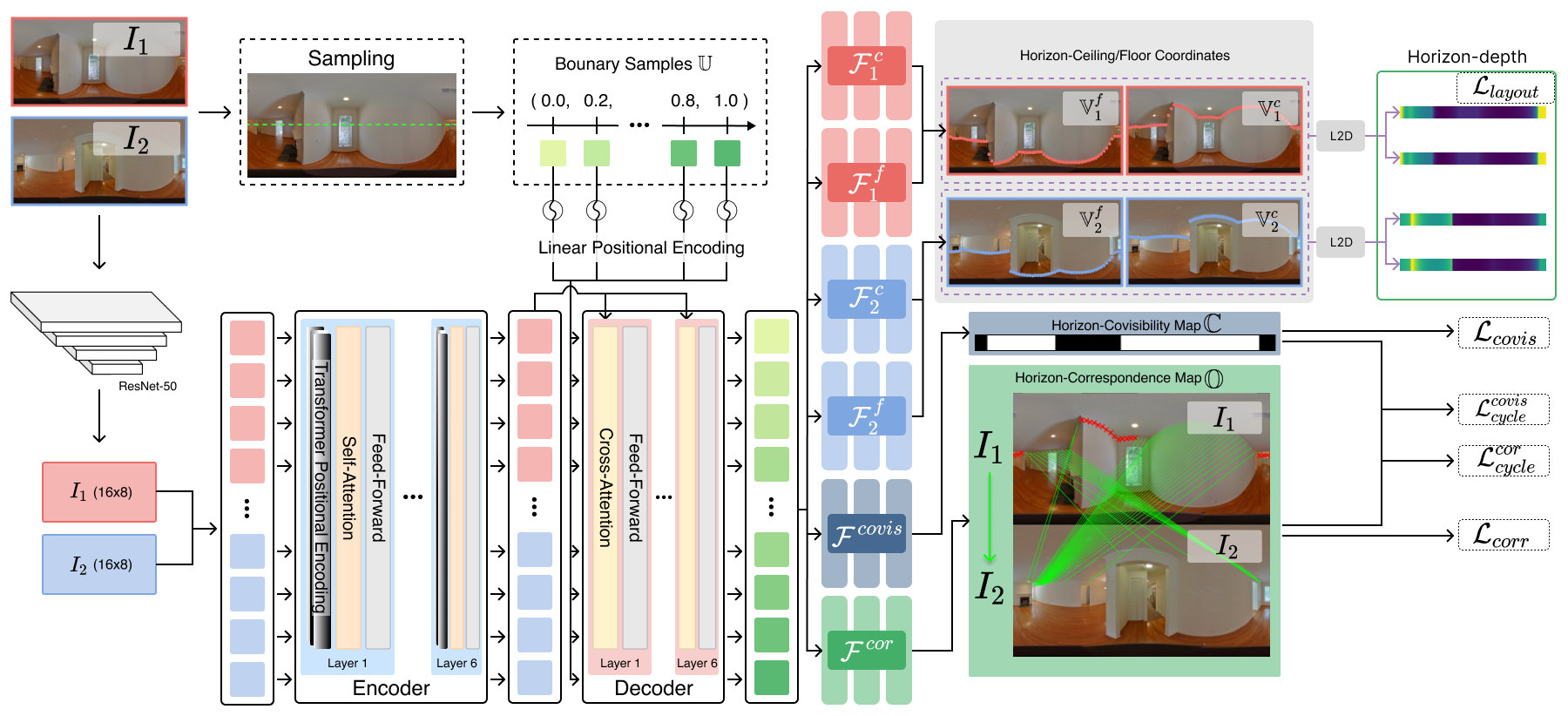}
    
    %\put(38.5, 24.5){\begin{minipage}{0.25\textwidth}\centering Conditional \\ Image \\ Generator  ${G_\theta}$ \end{minipage}}
    %\put(39, 7){\begin{minipage}{0.25\textwidth}\centering  Appearance \\ Field  ${F_\theta}$ \end{minipage}}
    %\put(55, 15.5){\begin{minipage}{0.5\textwidth}\centering ${L_{1}} + {L_{perceptual}} + {L_{gradient}}$\end{minipage}}
    % + Gradient smooth loss \\  (\secref{loss_function})\end{minipage}}
    %\put(3.5, 19){\begin{math} (D_1, S_1) \end{math}}
    %\put(15.5, 19){\begin{math} (D_2, S_2) \end{math}}
    %\put(24.5, 19){....}
    %\put(28.8, 19){\begin{math} (D_v, S_v) \end{math}}
    %\put(64.3, 20.5){\begin{math} G(D_1, S_1) \end{math}}
    %\put(76.5, 20.5){\begin{math} G(D_2, S_2) \end{math}}
    %\put(70, 18.7){Photorealistic panoramic images}
    %\put(86.7, 20.5){...}
    %\put(90.5, 20.5){\begin{math} G(D_v, S_v) \end{math}}
    %\put(31.2, 11.5){ $\appearanceembed$ \begin{math}\in \mathbb{R}^8 \end{math}}
    %\put(31, 7.3){\begin{math} s \in [1, k] \end{math}}
    %\put(30.5, 3){\begin{math} r(X, Y, Z) \end{math}}
    % \put(9.5, 0){\begin{minipage}{0.5\textwidth}\centering (Section \ref{sec:field_representation}, \ref{sec:appearance_embedding}) \end{minipage}}
    %\put(4.7, -1.8){\begin{minipage}{0.25\textwidth}\centering Input Scene \end{minipage}}
    %\put(68, 1){\begin{minipage}{0.25\textwidth}\centering Synthesized results \end{minipage}}
    \end{overpic}
    \\
    \caption{\tb{{\networkName} architecture.} Our network architecture follows the standard vision transformer with an encoder-decoder scheme and multiple MLP heads. Given a pair of panoramas $\panofirst$ and $\panosecond$, our network first extracts feature maps as input tokens by ResNet-50 and feeds those features into the transformer $\transformer$. We then treat the output tokens as boundary samples $\panouinputset$ in panorama $\panofirst$ and process the output of the transformer $\transformer$ by different MLP heads for horizon ceiling/floor coordinates, namely $\layoutset{k=(1,2)}{j=(c,f)}$, horizon-covisibility map $\covisoutputset$, and horizon-correspondence map $\coroutputset$, respectively. We perform L2D transformation on horizon ceiling/floor coordinates $\layoutset{k=(1,2)}{j=(c,f)}$ to obtain horizon-depth map $\mathbb{D}$. Finally, we compute the layout loss $\losslayout$, covisibility loss $\losscovisibility$, correspondence loss $\losscorrespondence$, and cycle-consistentcy loss $\losscycleconsistency$ using the corresponding 1D output horizon maps.
    }
\label{fig:network_archi}
\end{figure*}

\subsection{Network architecture}
\label{sec:network_archi}

\Figref{fig:network_archi} illustrates the {\networkName} architecture. \networkName{} uses building blocks from COTR~\citep{Jiang_2021_ICCV} and LED$^2$-Net~\citep{Wang_2021_CVPR}.
First, we feed two (vertically) axis aligned panoramas $\panofirst$ and $\panosecond$ into a ResNet-50~\citep{He2015} feature extractor and generate two feature maps of resolution $16\times8$. % We then concatenate these two feature maps \textit{side-by-side}, generating a single feature map of size $16\times16$.
Following the encoder-decoder transformer architecture, we feed the extracted feature maps into the transformer encoder block using the two sets of $16\times8$ pixels as input tokens. The output tokens of the transformer encoder block will be used for cross attention in the transformer decoder block.
We use a 2D $UV$ coordinate system to parametrize a panorama image with $(u,v)$ in $[0,1] \times [-1,1]$. The $u$ coordinate describes the horizontal position and the $v$ coordinate the vertical position.
We want to query multiple values (height, depth, correspondences, covisibility) for different $u$ coordinates.
We therefore uniformly sample the $u$ coordinate $\in [0,1]$ with $N$ evenly distributed samples. We obtain the set $\panouinputset= \{ \panouinput \}_{i=1}^N$. 
The query samples are encoded by linear positional encoding and are the input tokens of the transformer decoder.
The output tokens of the transformer decoder encode multiple types of information about the boundary samples. 
To extract this information, the output tokens are further processed using multiple MLP heads.
Specifically, we use:
(i) \emph{layout MLP heads} $\mlplayoutceiling{k}$ and $\mlplayoutfloor{k}$ with $k=(1,2)$. The outputs of $\mlplayoutceiling{k}$ and $\mlplayoutfloor{k}$ are the $v$ coordinates of the ceiling and floor boundaries in the panorama images, respectively.
%
%$\layoutset{k=(1,2)}{c} = \{ \layoutcoord{c} \}_{i=1}^N$, and horizon-floor coordinates, $\layoutset{k=(1,2)}{f} = \{ \layoutcoord{f} \}_{i=1}^N$.
We denote the $v$ coordinates of the ceiling boundaries as $\layoutset{k=(1,2)}{c}$ and the floor boundaries, $\layoutset{k=(1,2)}{f}$.
For each $\layoutset{k=(1,2)}{j=(c,f)}$, we further exploit the Layout-to-Depth (L2D) transformation~\citep{Wang_2021_CVPR} to generate a corresponding horizon-depth map $\horizondepthset{k=(1,2)}{j=(c,f)}$;
(ii) \emph{correspondence MLP head} $\mlpcoroutputset$ that outputs a horizon-correspondence map $\coroutputset = \{ \coroutput \}_{i=1}^N$, where $\coroutput$ indicates the correspondence between $\panouinput \in \panofirst$ and $\coroutput \in \panosecond$; and
(iii) \emph{covisibility MLP head} $\mlpcovisoutputset$ that outputs a horizon-covisibility map $\covisoutputset = \{ \covisoutput \}_{i=1}^N$, where $\covisoutput$ is a value $\in [0,1]$, encoding whether the $i-th$ element in $\coroutputset$ should be considered ($\covisoutput=1$) or not ($\covisoutput=0$) in the pose estimation.

\subsection{Loss functions}
\label{sec:loss_functions}
Here we elaborate on the layout, correspondence, covisibility, and cycle-consistency loss functions used for training our network.

\heading{Layout loss} calculates the low-level geometry loss between the predicted horizon-depth maps $\horizondepthset{}{}$ of input panoramas $\panofirst$ and $\panosecond$ and their ground-truth horizon-depth maps $\horizondepthsetgt{}{}$. The loss is defined as follows:
\begin{equation}
\label{eq:layout_loss}
    \losslayout = \frac{1}{M} \sum_{j=(c,f)}\sum_{k=(1,2)} \| \horizondepthset{k}{j} - \horizondepthsetgt{k}{j} \|_1,
\end{equation}
% where $M$ is the number of boundary samples.
where $M$ is the dimension of the horizon-depth maps.
%\begin{equation}
%    \losslayout = \frac{1}{M} \sum \| \horizondepthfirstceiling - \horizondepthfirstgt \|_1 + \| \horizondepthfirstfloor - \horizondepthfirstgt \|_1 + \| \horizondepthsecondceiling - \horizondepthsecondgt \|_1 + \| \horizondepthsecondfloor - \horizondepthsecondgt \|_1
%\end{equation}
%where the two floor view horizon-depth maps $\horizondepthfirstfloor$ and $\horizondepthsecondfloor$ come from the L2D transformation~\citep{Wang_2021_CVPR} on the layout floor boundary point sets $\layoutfirstfloorset$ and $\layoutsecondfloorset$. As for the two ground-truth horizon-depth maps $\horizondepthfirstgt$ and $\horizondepthsecondgt$, we derive them from the ground-truth layout.

\heading{Covisibility loss} evaluates the predicted normalized horizon co-visibility map $\covisoutputset = \{ \covisoutput \}_{i=1}^N $ with respect to the ground-truth horizon co-visibility map $\covisoutputgtset = \{ \covisoutputgt \}_{i=1}^N$. The loss is defined as follows:
\begin{equation}
\label{eq:co-visibility_loss}
    \losscovisibility = \frac{1}{N} \sum_{i=1}^N \alpha \covisoutputgt \cdot \log (\covisoutput) + (1 - \covisoutputgt) \cdot \log (1-\covisoutput) ,
\end{equation}
where $\alpha$ is the hyperparameter for weighting the positive samples. We use $\alpha$ because there are a lot more positive than negative samples in the ground truth.

\heading{Correspondence loss} calculates the difference between the predicted horizon-correspondence map $\coroutputset = \{ \coroutput \}_{i=1}^N$ and the ground-truth horizon-correspondence map $\coroutputgtset = \{ \coroutputgt \}_{i=1}^N$.
%, where $\{ \coroutput \}_{i=1}^N = \mlpcoroutputset(\transformer(\{ \panouinput \}_{i=1}^N))$.
The loss is defined as follows:
\begin{equation}
\label{eq:correspondence_loss}
    \losscorrespondence = \frac{1}{N} \sum_{i=1}^N
    \left \{ 
    \begin{array}{rcl}
        \min(\| \coroutput - \coroutputgt \|_1 , \| 1 - \coroutputgt + \coroutput \|_1), & \text{if } \covisoutputgt \geq 0.5\\
        0, & \text{otherwise}
    \end{array} \right. ,
\end{equation}
where we use a cyclic loss instead of the simple L1 loss between the predicted and ground-truth correspondence to adopt the coordinate system of in equirectangular projection.
% \james{I don't understand this...}

\heading{Cycle-consistency loss~\citep{Jiang_2021_ICCV}} enforces the network outputs to be cycle-consistent and adapts the network to different ray casting positions in contrast to the uniformly sampled ray casting positions. We reverse the order of the two panoramas $\panofirst$ and $\panosecond$, and treat the ground-truth horizon-correspondence map $\{ \coroutputgt \}_{i=1}^N$ as input and the original input $\{ \coroutput \}_{i=1}^N$ as target correspondence. We separate the cycle-consistency loss into two parts as follows:
\begin{equation}
\label{eq:cyclecorr_loss}
    \losscyclecorr = \frac{1}{N} \sum_{i=1}^N
    \left \{ 
    \begin{array}{rcl}
        \min(\| \mlpcoroutputset(\transformer(\coroutputgt)) - \panouinput \|_1 , \| 1 - \panouinput + \mlpcoroutputset(\transformer(\coroutputgt)) \|_1), & \text{if } \covisoutputgt \geq 0.5\\
        0, & \text{otherwise}
    \end{array} \right.
\end{equation}
and
\begin{equation}
\label{eq:cyclecovis_loss}
    \losscyclecovis = \frac{1}{N} \sum_{i=1}^N
    \alpha \covisoutputgt \cdot \log (\mlpcovisoutputset(\transformer(\coroutputgt))) + (1 - \covisoutputgt) \cdot \log (1-\mlpcovisoutputset(\transformer(\coroutputgt))) .
\end{equation}
Here, we still use the ground-truth horizon-covisibility map $\{ \covisoutputgt \}_{i=1}^N$ as the target in $\losscyclecovis$ since the order of $\{ \coroutputgt \}_{i=1}^N$ and $\{ \coroutput \}_{i=1}^N$ is the same.
%
% The $\losscycleconsistency$ is thus defined as follows:

% \begin{equation}
%     \losscycleconsistency = \losscyclecovis + \losscyclecorr
% \end{equation}

Finally, the overall loss function used in our network is defined as follows:
\begin{equation}
\label{eq:total_loss}
    \losstotal = \lambda_1 \losslayout + \lambda_2 \losscorrespondence + \lambda_3 \losscovisibility + \lambda_4 \losscyclecorr + \lambda_5 \losscyclecovis,
\end{equation}
where $\lambda_1$, $\lambda_2$, $\lambda_3$, $\lambda_4$, and $\lambda_5$ are the hyperparameters for weighting the loss functions.

\begin{figure*}[!t]
    % \begin{subfigure}[!t]{\linewidth}
    %   \includegraphics[width=\linewidth,keepaspectratio]{figure/images/overview.jpg}
    % \end{subfigure}
    % \begin{overpic}[width=\textwidth]{iclr2023/Figures/images/Registration_onerow.png}
    % \begin{overpic}[width=\textwidth]{iclr2023/Figures/images/Registration_wray_two.png}
    \begin{overpic}[width=\textwidth]{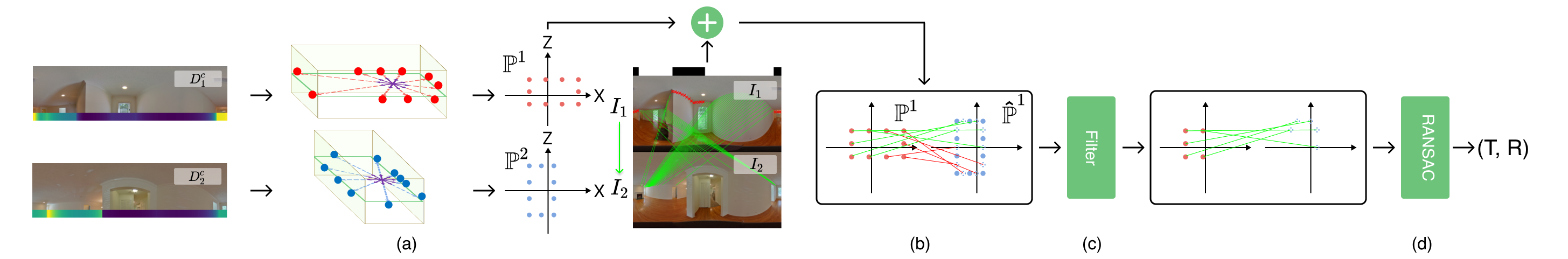}
    
    %\put(38.5, 24.5){\begin{minipage}{0.25\textwidth}\centering Conditional \\ Image \\ Generator  ${G_\theta}$ \end{minipage}}
    %\put(39, 7){\begin{minipage}{0.25\textwidth}\centering  Appearance \\ Field  ${F_\theta}$ \end{minipage}}
    %\put(55, 15.5){\begin{minipage}{0.5\textwidth}\centering ${L_{1}} + {L_{perceptual}} + {L_{gradient}}$\end{minipage}}
    % + Gradient smooth loss \\  (\secref{loss_function})\end{minipage}}
    %\put(3.5, 19){\begin{math} (D_1, S_1) \end{math}}
    %\put(15.5, 19){\begin{math} (D_2, S_2) \end{math}}
    %\put(24.5, 19){....}
    %\put(28.8, 19){\begin{math} (D_v, S_v) \end{math}}
    %\put(64.3, 20.5){\begin{math} G(D_1, S_1) \end{math}}
    %\put(76.5, 20.5){\begin{math} G(D_2, S_2) \end{math}}
    %\put(70, 18.7){Photorealistic panoramic images}
    %\put(86.7, 20.5){...}
    %\put(90.5, 20.5){\begin{math} G(D_v, S_v) \end{math}}
    %\put(31.2, 11.5){ $\appearanceembed$ \begin{math}\in \mathbb{R}^8 \end{math}}
    %\put(31, 7.3){\begin{math} s \in [1, k] \end{math}}
    %\put(30.5, 3){\begin{math} r(X, Y, Z) \end{math}}
    % \put(9.5, 0){\begin{minipage}{0.5\textwidth}\centering (Section \ref{sec:field_representation}, \ref{sec:appearance_embedding}) \end{minipage}}
    %\put(4.7, -1.8){\begin{minipage}{0.25\textwidth}\centering Input Scene \end{minipage}}
    %\put(68, 1){\begin{minipage}{0.25\textwidth}\centering Synthesized results \end{minipage}}
    \end{overpic}
    \\
    % \caption{\tb{Registration pipeline}. We start our registration process from the two horizon-depth maps $\horizondepthset{1}{c}$ and $\horizondepthset{2}{c}$. (a) They are projected to the XZ plane to become 2D boundary point sets $\layoutfirstpointset$ and $\layoutsecondpointset$. (b) Considering the non-uniformly distributed horizon-correspondence map $\coroutputset$, we extract the one-to-one point-wise correspondence $\layoutfirstpointset$ and $\layoutfirstpointsetmatch$ by interpolating $\layoutsecondpointset$ with $\coroutputset$. Finally, (c) we filter out some of the in-covisible pairs via horizon co-visibility map $\covisoutputset$ and (d) robustly estimate final pose parameters by RANSAC.    }
    \caption{\tb{Registration pipeline}. We start our registration process by (a) utilizing the elementary geometric transformations~\citep{Wang_2021_CVPR} to obtain: 1) two horizon-depth maps $\horizondepthset{1}{c}$ and $\horizondepthset{2}{c}$, and 2) two 2D point sets $\layoutfirstpointset$ and $\layoutsecondpointset$. (b) Considering the non-uniformly distributed horizon-correspondence map $\coroutputset$, we extract the one-to-one point-wise correspondence $\layoutfirstpointset$ and $\layoutfirstpointsetmatch$ by interpolating $\layoutsecondpointset$ with $\coroutputset$. Finally, (c) we filter out some of the in-covisible pairs via horizon co-visibility map $\covisoutputset$ and (d) robustly estimate final pose parameters by RANSAC.    }
\label{fig:registration_archi}
\end{figure*}

%\subsection{Registration}
\subsection{Non-linear registration}
\label{sec:registration}
\Figref{fig:registration_archi} illustrates the pipeline of our non-linear registration.
%
%Given two estimated upper (ceiling) horizon-depth maps $\horizondepthset{1}{c}$ and $\horizondepthset{2}{c}$, we multiply individual horizon-depth maps element-wise with a uniformly-sampled unit sphere to obtain two sets of boundary points $\layoutfirstpointset = \{ \layoutfirstpoint \}_{i=1}^M$ and $\layoutsecondpointset = \{ \layoutsecondpoint \}_{i=1}^M$ on the layouts, where $M$ is the number of samples on the unit sphere (\Figref{fig:registration_archi}(a)). 
% Given two estimated upper (ceiling) horizon-depth maps $\horizondepthset{1}{c}$ and $\horizondepthset{2}{c}$, we project each of them to the XZ plane to obtain two sets of 2D boundary points $\layoutfirstpointset = \{ \layoutfirstpoint \}_{i=1}^M$ and $\layoutsecondpointset = \{ \layoutsecondpoint \}_{i=1}^M$ on the layouts, where $M$ is the number of samples on the unit sphere (\Figref{fig:registration_archi}(a)).
%
We use a 3D Cartesian coordinate system to perform our registration process where the $y$-axis is the up axis, the camera center is the origin, and the $XZ$-plane is parallel to floor and ceiling.
%
% First, our goal is to obtain two sets of boundary points $\layoutfirstpointset$ and $\layoutsecondpointset$ on the layouts to perform the registration. 
We reuse the elementary geometric transformations described by~\cite{Wang_2021_CVPR} to obtain the following:
1) the horizon-depth maps $\horizondepthset{1}{c} = \{ d_i^1 \in \mathbb{R}^1 \}_{i=1}^M$ and $\horizondepthset{2}{c} = \{ d_i^2 \in \mathbb{R}^1 \}_{i=1}^M$, which are derived from the predicted ceiling layout boundaries $\layoutset{}{c}$ and represent the $M$ evenly sampled distances from the origin to the predicted layout boundary in the $XZ$-plane; 
%
%2) the horizon-depth maps $\horizondepthset{1}{f}$ and $\horizondepthset{2}{f}$ derived from $\layoutset{}{f}$ analogously and
%
2) two sets of 2D points $\layoutfirstpointset = \{ \layoutfirstpoint \in \mathbb{R}^2 \}_{i=1}^M$ and $\layoutsecondpointset = \{ \layoutsecondpoint \in \mathbb{R}^2 \}_{i=1}^M$. These points are on the layout boundary on the $XZ$-plane. Each point corresponds to a depth value in $\horizondepthset{1}{c}$ or $\horizondepthset{2}{c}$, respectively (\Figref{fig:registration_archi}(a)). 
%
% Given a unit vector set $\mathcal{S}^2 = \{\vec{n_i} \in \mathbb{R}^3 : \|\vec{n_i}\| =1 \}_{i=1}^M$ sampled uniformly on the $XZ$-plane, we multiply individual horizon-depth maps element-wise with the unit vector set $\mathcal{S}^2$ and obtain 
%
We would like to remark that we only consider 3-DoF transformations in ZInD~\citep{Cruz_2021_CVPR} following \citep{Wang_2022_CVPR} and that the conversion of~\cite{Wang_2021_CVPR} includes a resampling step from $N$ to $M$ boundary samples.
In order to compute a one-to-one point-wise correspondence between $\layoutfirstpointset$ and $\layoutsecondpointset$, we exploit the estimated horizon-correspondence map $\coroutputset$. 
Since $\coroutputset$ is not necessary a uniform distribution, for each $\layoutfirstpoint \in \layoutfirstpointset$, we compute its corresponding point via interpolating $\layoutsecondpointset$ with $\coroutputset$ (\Figref{fig:registration_archi}(b)).
We denote the corresponding boundary points as $\layoutfirstpointsetmatch = \{ \layoutfirstpointmatch \}_{i=1}^M$.
We then filter out the matched pair in $\layoutfirstpointset$ and $\layoutfirstpointsetmatch$ according to the horizon co-visibility map $\covisoutputset$ (\Figref{fig:registration_archi}(c)).
The final pose parameters (\ie translation $\translation$ and rotation $\rotation$) are computed via a RANSAC-based estimation (Figure 3(d)).

\subsection{Layout fusion}
\label{sec:fusion}
Given the relative camera pose between input panoramas, we combine two individual partial layouts into a unified one as follows.
First, for each input panorama, we adopt the same post-processing procedure as LED$^2$-Net~\citep{Wang_2021_CVPR} that converts the estimated horizon-ceiling coordinates $\layoutset{k}{c}$ and horizon-floor coordinates $\layoutset{k}{f}$ into a 2D layout map $L_{k}$ and a layout height $H_{k}$.
Then, we register two 2D layout maps using the estimated relative camera pose and then combine them into a complete 2D layout map via a union operation $L_{final}=L_{1} \cup L_{2}$.
The final 3D layout is obtained by extruding $L_{final}$ with the average layout height $(H_1+H_2)/2$.

\if 1
\subsection{Problem formulation}
Let $\panouinputset= \{ \panouinput \}_{i=1}^N$ be the normalized horizontal coordinate set of the query rays in panorama $\panofirst$, for which we wish to find the ceiling vertical layout coordinate set~\citep{Wang_2021_CVPR} $\layoutfirstceilingset = \{ \layoutfirstceiling \}_{i=1}^N$ and floor coordinate set $\layoutfirstfloorset = \{ \layoutfirstfloor \}_{i=1}^N$ in panorama $\panofirst$, $\layoutsecondceilingset = \{ \layoutsecondceiling \}_{i=1}^N$ and $\layoutsecondfloorset = \{ \layoutsecondfloor \}_{i=1}^N$ in panorama $\panosecond$, horizon-correspondence map $\coroutputset = \{ \coroutput \}_{i=1}^N$ between panoramas $\panofirst$ and $\panosecond$, and horizon co-visibility map $\covisoutputset = \{ \covisoutput \}_{i=1}^N$.
% by COTR
We treat our goal of predicting layout coordinate sets, horizon-correspondence map, and horizon co-visibility map as to find the best set of parameter $\transformerparam$, $\mlpfirstceilingparam$, $\mlpfirstfloorparam$, $\mlpsecondceilingparam$, $\mlpsecondfloorparam$, $\mlpcoroutputsetparam$, and $\mlpcovisoutputsetparam$ in transformer $\transformer$, first ceiling layout MLP head $\mlpfirstceiling$, first floor layout MLP head $\mlpfirstfloor$, second ceiling layout MLP head $\mlpsecondceiling$, second floor layout MLP head $\mlpsecondfloor$, horizon-correspondence MLP head $\mlpcoroutputset$, and horizon co-visibility MLP head $\mlpcovisoutputset$ minimizing
\begin{equation}
    \mathop{\arg\min}_{\transformerparam , \mlpfirstceilingparam , \mlpfirstfloorparam , \mlpsecondceilingparam , \mlpsecondfloorparam, \mlpcoroutputsetparam , \mlpcovisoutputsetparam} \mathop{\mathbb{E}}_{(\panouinput, \layoutfirstceiling, \layoutfirstfloor, \layoutsecondceiling, \layoutsecondfloor, \coroutput, \covisoutput, \panofirst, \panosecond ) D} \lambda_1 \losslayout + \lambda_2 \losscorrespondence + \lambda_3 \losscovisibility + \lambda_4 \losscycleconsistency
\end{equation}
where $D$ is the training dataset of ground truth, $L_{layout}$ measures the layout loss inspired by~\citep{Wang_2021_CVPR}, $L_{corr}$ measures the correspondence prediction accuracy given $\panouinput$, $L_{covis}$ measures the accuracy of predicted co-visibility mask, and $L_{cycle}$ improves the cycle consistency of the correspondence and layout prediction. Please refer to the \secref{sec:loss_functions} for the details of all the loss functions.

\subsection{Network architecture}

The overall architecture overview is illustrated in \figref{fig:overview}.
We use building blocks of the architectures COTR~\citep{Jiang_2021_ICCV} and LED$^2$Net~\citep{Wang_2021_CVPR} to construct our network.
First, we feed both (vertically) axis aligned~\citep{Wang_2022_CVPR} panoramas $\panofirst$ and $\panosecond$ into a ResNet-50 feature extractor and generate two feature maps of size \eric{16x8}.
We then concatenate these two feature maps \textit{side-by-side}, generating a single feature map of size \eric{16x16}.
Following the vision transformer architecture~\citep{Jiang_2021_ICCV}, we feed the extracted feature maps into the transformer $\transformer$, interpreting the results with the input query ray $\panouinputset$ which encoded by linear positional encoding~\citep{Jiang_2021_ICCV}.
Finally, we process the output of the transformer $\transformer$ using multiple MLP heads: first ceiling layout MLP head $\mlpfirstceiling$, first floor layout MLP head $\mlpfirstfloor$, second ceiling layout MLP head $\mlpsecondceiling$, second floor layout MLP head $\mlpsecondfloor$, horizon-correspondence MLP head $\mlpcoroutputset$, and horizon co-visibility MLP head $\mlpcovisoutputset$. Thus we can obtain four layout coordinate sets $\layoutfirstceilingset$, $\layoutfirstfloorset$, $\layoutsecondceilingset$, $\layoutsecondfloorset$, horizon-correspondence map $\coroutputset$, and horizon co-visibility map $\covisoutputset$, respectively.
\eric{Please refer to the supplementary for more details about our network architecture.}

\subsection{Loss functions}
\label{sec:loss_functions}
Here we elaborate on the layout, correspondence, and co-visibility loss functions and the cycle-consistency used for training our network.

\heading{Layout loss} calculates the low-level geometry loss between the predicted and ground-truth horizon-depth maps. We follow LED$^2$-Net using the L1 loss to measure the errors:
\begin{equation}
    \losslayout = \frac{1}{M} \sum \| \horizondepthfirstceiling - \horizondepthfirstgt \|_1 + \| \horizondepthfirstfloor - \horizondepthfirstgt \|_1 + \| \horizondepthsecondceiling - \horizondepthsecondgt \|_1 + \| \horizondepthsecondfloor - \horizondepthsecondgt \|_1
\end{equation}
where the two floor view horizon-depth maps $\horizondepthfirstfloor$ and $\horizondepthsecondfloor$ come from the L2D transformation~\citep{Wang_2021_CVPR} on the layout floor boundary point sets $\layoutfirstfloorset$ and $\layoutsecondfloorset$. As for the two ground-truth horizon-depth maps $\horizondepthfirstgt$ and $\horizondepthsecondgt$, we derive them from the ground-truth layout.

\heading{Co-visibility loss} evaluates the predicted normalized horizon co-visibility map $\covisoutputset = \{ \covisoutput \}_{i=1}^N \in [0, 1]$ with respect to the ground-truth horizon co-visibility map $\covisoutputgtset = \{ \covisoutputgt \}_{i=1}^N \in [0, 1]$, where $\{ \covisoutput \}_{i=1}^N = \mlpcovisoutputset(\transformer(\{ \panouinput \}_{i=1}^N)) $. The co-visibility loss is defined as follows:
\begin{equation}
    \losscovisibility = \frac{1}{N} \sum_{i=1}^N \alpha \covisoutputgt \cdot \log (\covisoutput) + (1 - \covisoutputgt) \cdot \log (1-\covisoutput) ,
\end{equation}
where $\alpha$ is the hyperparameter for weighting the positive samples.

\heading{Correspondence loss} calculates the predicted horizon-correspondence map $\coroutputset = \{ \coroutput \}_{i=1}^N$ and the ground-truth horizon-correspondence map $\coroutputgtset = \{ \coroutputgt \}_{i=1}^N$, where $\{ \coroutput \}_{i=1}^N = \mlpcoroutputset(\transformer(\{ \panouinput \}_{i=1}^N))$. The correspondence loss is defined as follows:
\begin{equation}
    \losscorrespondence = \frac{1}{N} \sum_{i=1}^N
    \left \{ 
    \begin{array}{rcl}
        \min(\| \coroutput - \coroutputgt \|_1 , \| 1 - \coroutputgt + \coroutput \|_1), & \text{if } \covisoutputgt \geq 0.5\\
        0, & \text{otherwise}
    \end{array} \right. ,
\end{equation}
where we use a cyclic loss instead of the simple l1 loss between the predicted and ground-truth correspondence, in order to adopt the coordinate system of the equirectangular projection.

\heading{Cycle-consistency loss~\citep{Jiang_2021_ICCV}} enforces the network outputs to be cycle-consistent and adapts the network to different ray casting positions in contrast to the uniformly sampled ray casting positions. We reverse the order of the two panoramas $\panofirst$ and $\panosecond$, and treat the ground-truth horizon-correspondence map $\{ \coroutputgt \}_{i=1}^N$ as input and the original input $\{ \coroutput \}_{i=1}^N$ as target correspondence. We separate the cycle-consistency loss into two parts as follows:
\begin{equation}
    \losscyclecovis = \frac{1}{N} \sum_{i=1}^N
    \alpha \covisoutputgt \cdot \log (\mlpcovisoutputset(\transformer(\coroutputgt))) + (1 - \covisoutputgt) \cdot \log (1-\mlpcovisoutputset(\transformer(\coroutputgt)))
\end{equation}
and
\begin{equation}
    \losscyclecorr = \frac{1}{N} \sum_{i=1}^N
    \left \{ 
    \begin{array}{rcl}
        \min(\| \mlpcoroutputset(\transformer(\coroutputgt)) - \panouinput \|_1 , \| 1 - \panouinput + \mlpcoroutputset(\transformer(\coroutputgt)) \|_1), & \text{if } \covisoutputgt \geq 0.5\\
        0, & \text{otherwise}
    \end{array} \right.
\end{equation}
where we still use the ground-truth horizon co-visibility map $\{ \covisoutputgt \}_{i=1}^N$ as the target in $\losscyclecovis$ since the order of $\{ \coroutputgt \}_{i=1}^N$ and $\{ \coroutput \}_{i=1}^N$ is the same.
The $\losscycleconsistency$ is defined as follows:

\begin{equation}
    \losscycleconsistency = \losscyclecovis + \losscyclecorr
\end{equation}

Finally, the overall loss function used in our network is defined as follows:
\begin{equation}
    \losstotal = \lambda_1 \losslayout + \lambda_2 \losscorrespondence + \lambda_3 \losscovisibility + \lambda_4 \losscycleconsistency,
\end{equation}
where $\lambda_1$, $\lambda_2$, $\lambda_3$, and $\lambda_4$ are the hyperparameters for weighting the loss functions.

%\subsection{Registration}
\subsection{Non-linear registration}
\Figref{fig:registration_archi} illustrates the pipeline of our non-linear registration process.
Given the predicted results including two predicted ceiling vertical layout coordinate sets $\layoutfirstceilingset$ and $\layoutsecondceilingset$, horizon-correspondence map $\coroutputset$, and horizon co-visibility map $\covisoutputset$, the relative camera pose is computed as follows.
We first derive the two ceiling view horizon-depth map $\horizondepthfirstceiling$ and $\horizondepthsecondceiling$ by feeding the two predicted ceiling vertical layout coordinate set $\layoutfirstceilingset$ and $\layoutsecondceilingset$. Next, the two sets are sent to L2D~\citep{Wang_2021_CVPR} to compute transformations $\horizondepthfirstceiling = L2D(\layoutfirstceilingset)$ and $\horizondepthsecondceiling = L2D(\layoutsecondceilingset)$.
We then perform element-by-element multiplication of these two horizon-depth maps $\horizondepthfirstceiling$, $\horizondepthsecondceiling$ and the uniformly-sampled unit sphere. Afterwards, each panorama's layout point set $\layoutfirstpointset = \{ \layoutfirstpoint \}_{i=1}^M$ and $\layoutsecondpointset = \{ \layoutsecondpoint \}_{i=1}^M$ are computed. Note that the point sets are strictly on the XZ plane since we only consider the 3 DoF transformations in ZInD~\citep{Cruz_2021_CVPR}.
In order to incorporate the horizon-correspondence map $\coroutputset$ with the two layout point sets $\layoutfirstpointset$ and $\layoutsecondpointset$, the order of the corresponding query ray $\panouinput$ of the layout point set $\layoutfirstpointset$ is the same with the horizon-correspondence map $\coroutputset$.
As our goal is to find the transformation using correspondence and layout point set, we treat two layout point sets $\layoutfirstpointset$ and $\layoutsecondpointset$ as the feature points. We then convert the horizon-correspondence map $\coroutputset$ into the correspondence between two feature point sets.
However, as we want to use RANSAC to find the final affine transformation between the two camera locations of the two panorama, the exact corresponding feature points' coordinate related to the $\layoutsecondpointset$ is still missing.
To this end, we leverage the horizon-correspondence map $\coroutputset$ to obtain the exact corresponding feature points' coordinates related to the $\layoutsecondpointset$.
Specifically, we apply the grid sample operation which computes the warped layout point set $\layoutsecondpointsetwarp$ using original layout point set $\layoutsecondpointset$ and locations from horizon-correspondence map $\coroutputset$.
We then filter out the feature-correspondence pair in $\layoutfirstpointset$ and $\layoutsecondpointsetwarp$ based on the horizon co-visibility map $\covisoutputset$.
Finally, we obtain the predicted translation $\translation$ and rotation $\rotation$ by feeding the filtered $\layoutfirstpointset$, $\layoutsecondpointsetwarp$ and their correspondences into RANSAC.
\fi
% \clearpage
\section{Results}
\label{sec:results}

In this section, we compare our method with state-of-the-art layout reconstruction and panorama registration approaches. We also conducted multiple ablation studies to validate the necessity of individual modules in our architecture.

\subsection{Experimental Settings}
\heading{Dataset.}
We conduct all the experiments on a public indoor panorama dataset, Zillow Indoor Dataset (ZInD), which contains 67,448 indoor panoramas. We follow the same procedure as PSMNet to select the panorama pair instances and obtain training (105256), validation (12376), and tests (12918) pairs. While we tried to match the PSMNet test protocol as closely as possible and exchanges multiple emails with the authors to that effect, we are still waiting for the authors of PSMNet to release their testing code and dataset split.

\heading{Competing methods.}
We compare our method with the following state-of-the-art layout reconstruction models, LED$^2$-Net~\citep{Wang_2021_CVPR}, LGT-Net~\citep{Jiang_2022_CVPR}, and PSMNet~\citep{Wang_2022_CVPR}.
Since LED$^2$-Net and LGT-Net are single-view layout estimation methods, we first estimate the layout for each view, register two input panoramas using OpenMVG, and then perform a union operation to obtain the final reslut. 
Note that in cases where OpenMVG fails to produce a registration, we use average ground-truth pose of the training datasets.
To evaluate the performance on the panorama registration, we compare our {\networkName} with OpenMVG, a popular Structure-from-Motion library that supports stereo panorama matching. 
%We also compare the registration accuracy with OpenMVG, which is a state-of-the-art panorama registration method.
%
We followed the official settings for the feature extractor and correspondence matching and applied 'incrementalv2' mode for SfM operation to achieve the best reconstruction rate.
%First, we adopt the same settings of the feature extractor and correspondence matching strategy with the official spherical sample code.
%
%To achieve the higher reconstruction rate, we then applied 'incrementalv2' for SfM operation.

% Maybe also mention that OpenMVG had a recently improved version, what the settings were and how we went about finding the best settings for OpenMVG.

\heading{Evaluation metrics.}
We used 2D IoU, 3D IoU, and $\delta^i$ for quantitative evaluation of the layout reconstruction.
As for measuring the image registration quality, we used angular error between estimated relative pose and ground-truth in translation($\angularErr{t}$) and rotation($\angularErr{R}$)~\citep{Brachmann_2019_ICCV} and mean-average-accuracy~\citep{Jiang_2021_ICCV} of translation and rotation at a $5^\circ$ and $10^\circ$ error threshold, denoted as T-mAA$@5^{\circ}$, T-mAA$@10^{\circ}$, R-mAA$@5^{\circ}$, and R-mAA$@10^{\circ}$, respectively.

\heading{Implementation details.}
We implemented our model in PyTorch and conducted experiments on a single NVIDIA V100 with 32GB VRAM. The resolution of the panoramas is resized to $512 \times 256$. We use the Adam optimizer with b1=0.9 and b2=0.999. The learning rates of the transformer and the ResNet-50 are 1e-4 and 1e-5, and the batch size is set to 8. We empirically set $\lambda_1 = 1, \lambda_2 = 1, \lambda_3 = 1, \lambda_4 = 1, \lambda_5 = 1$ in \Eqref{eq:total_loss}, $\alpha = 0.1$ in \Eqref{eq:co-visibility_loss}, $M = 256$ in \Eqref{eq:layout_loss}, and $N = 256$ in \Secref{sec:loss_functions}.

\subsection{Evaluation on layout reconstruction}
\begin{table}[!t]
    \caption{\tb{Quantitative comparisons with state-of-the-art methods on layout reconstruction.} 
    The * symbol means that the numbers are reported in PSMNet~\cite{Wang_2022_CVPR}.
    %* means the methods test on the ZInD dataset with our testing set and the camera poses are derived from the OpenMVG or the average poses of the training dataset for the OpenMVG failure cases. The methods without * means that they're reported from PSMNet~\cite{Wang_2022_CVPR}.
    }
    \label{tab:result_layout_comp}
    \centering
    % \resizebox{\textwidth}{!}{
        \begin{tabular}{lllllll}
        \toprule
        & \multicolumn{3}{c}{w/ GT pose} & \multicolumn{3}{c}{w/o GT pose} \\
        \midrule
        Method & 2D IoU$\uparrow$ & $\delta^{i}$ $\uparrow$ & 3D IoU$\uparrow$ & 2D IoU$\uparrow$ & $\delta^i$ $\uparrow$ & 3D IoU$\uparrow$ \\ 
        \midrule
        %LED2Net~\cite{Wang_2021_CVPR} & 0.7639 & 0.9056 & - & 0.6581 & 0.8566 & - \\ 
        LED2Net~\citep{Wang_2021_CVPR} & 0.8364 & 0.9557 & 0.8131 & 0.5889 & 0.8777 & 0.5738 \\ 
        LGT-Net~\citep{Jiang_2022_CVPR} & 0.8388 & 0.9537 & 0.8126 & 0.5831 & 0.8779 & 0.5661 \\ 
        PSMNet*~\citep{Wang_2022_CVPR} & 0.8101 & 0.9238 & - & 0.7577 & 0.9217 & - \\ 
        \networkName~(Ours)    & \tb{0.8449} &\tb{0.9603} & \tb{0.8211} &\tb{0.8026} &\tb{0.9452} & \tb{0.7816} \\ 
        \bottomrule
        \end{tabular}
    % }
\end{table}

\begin{table}[!b]
    \caption{\textbf{Quantitative comparisons with state-of-the-arts methods on indoor panorama registration.} We evaluate all the methods on the ZInD dataset with our testing set.}
    \label{tab:result_reg_comp}
    \centering
    \resizebox{\textwidth}{!}{
        \begin{tabular}{lll|llll}
        \toprule
        & \multicolumn{2}{c}{SfM successful pair} & \multicolumn{4}{c}{full testing pair} \\
        \midrule
        Method & $\angularErr{t}\downarrow$ & $\angularErr{R}\downarrow$ &
        R-mAA$@5^{\circ} \uparrow$ & R-mAA$@10^{\circ} \uparrow$ & T-mAA$@5^{\circ} \uparrow$ & T-mAA$@10^{\circ} \uparrow$ \\ 
        \midrule
        OpenMVG~\citep{moulon2016openmvg}  & 17.6717 & 7.3707 & 0.1787 & 0.1969 & 0.1369 & 0.1627 \\ 
        % OpenMVG  & 12.2662 & 4.7936 & 0.2075 & 0.2146 & 0.1720 & 0.1928 \\ 
        \networkName~(Ours)    & \textbf{6.9907} &\textbf{2.5514} &\textbf{0.8633} &\textbf{0.9378} &\textbf{0.7667} &\textbf{0.8854}     \\ 
        \bottomrule
        \end{tabular}
    }
\end{table}

\begin{figure}[!t]
\centering
    \includegraphics[width=\columnwidth,keepaspectratio]{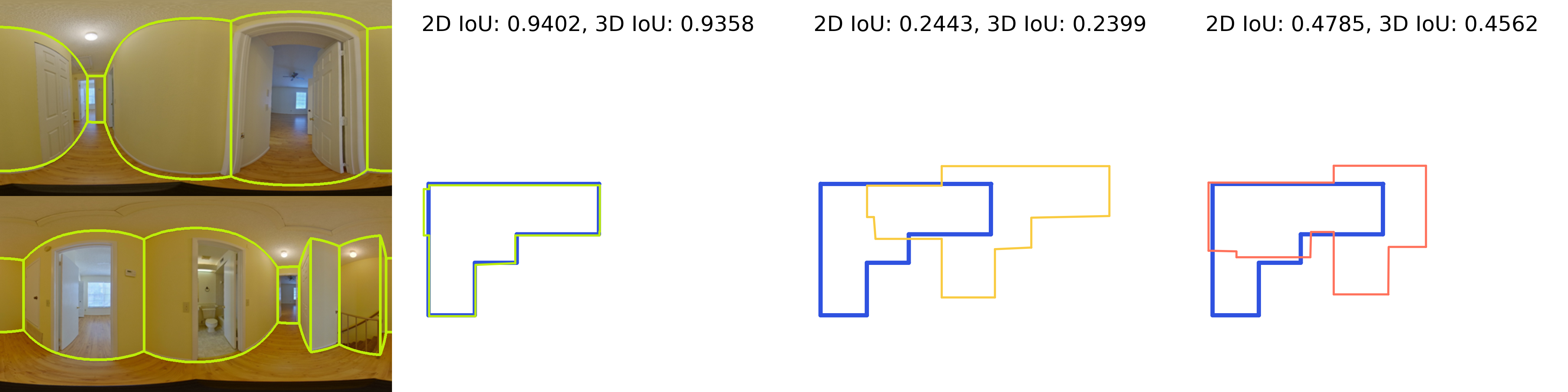}
    \includegraphics[width=\columnwidth,keepaspectratio]{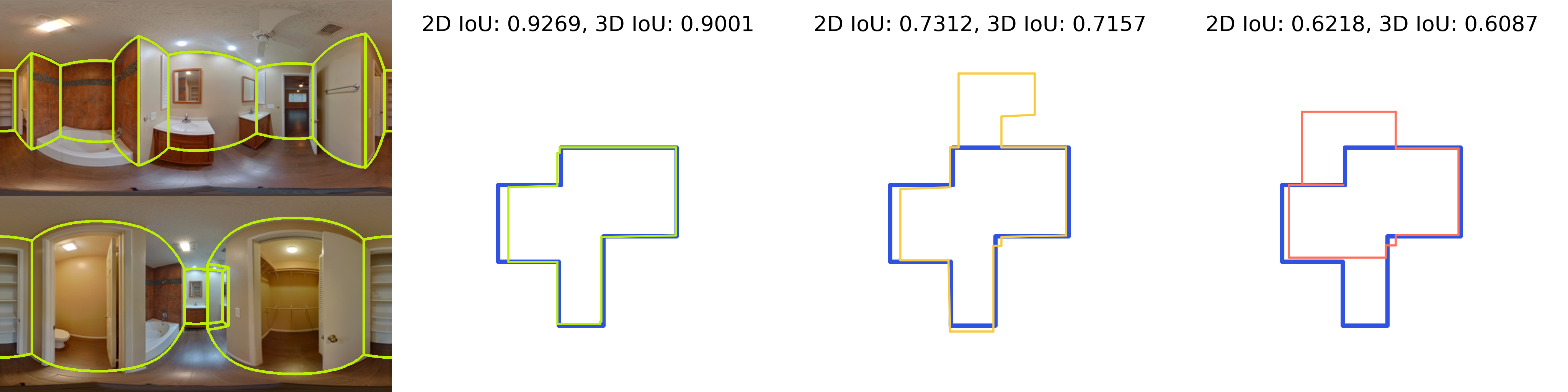}
    \includegraphics[width=\columnwidth,keepaspectratio]{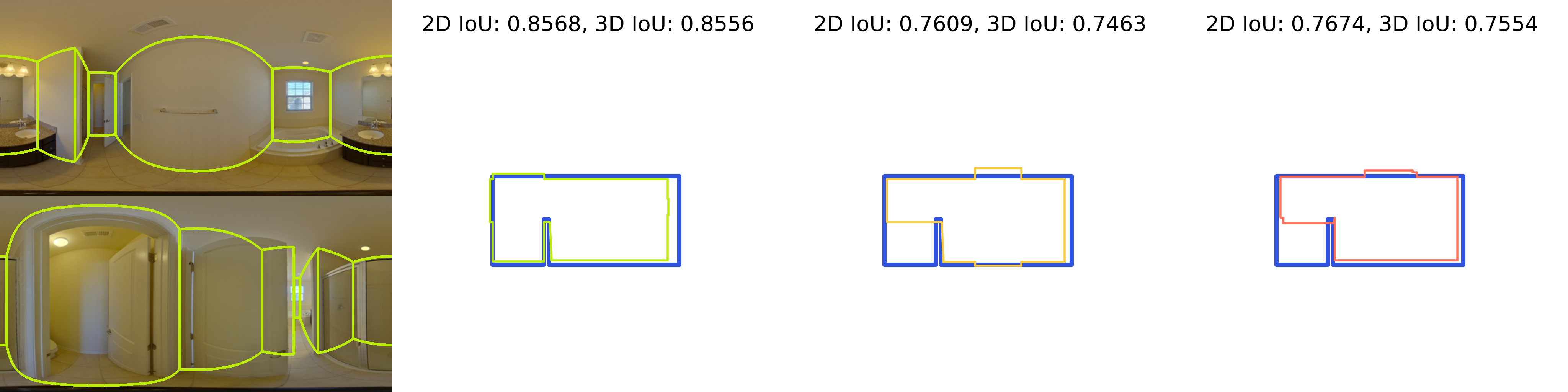}
    \includegraphics[width=\columnwidth,keepaspectratio]{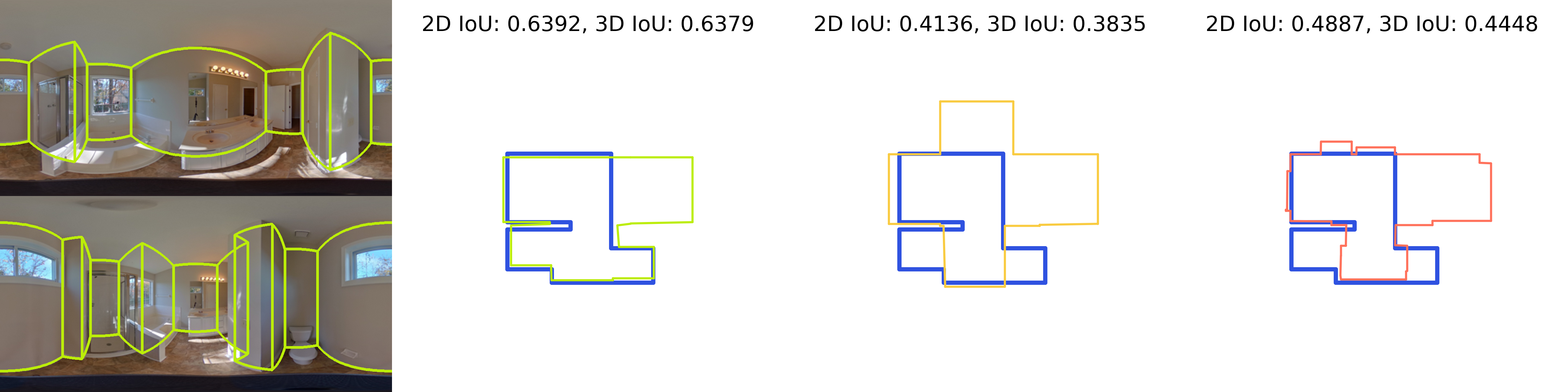}
    \begin{subfigure}[t]{.24\columnwidth}
    \caption*{Input}
    \end{subfigure}
    \begin{subfigure}[t]{.24\columnwidth}
    \caption*{Ours}
    \end{subfigure}
    \begin{subfigure}[t]{.24\columnwidth}
    \caption*{LED$^2$-Net}
    \end{subfigure}
    \begin{subfigure}[t]{.24\columnwidth}
    \caption*{LGT-Net}
    \end{subfigure}
    % \vspace{\figcapmargin}
    \caption{\textbf{Visual comparisons.} We show visual comparisons with other competing methods categorized by difficulty. From top to bottom, we select cases where our reconstruction accuracy is the range of top 10\%, 20\%, 50\%, and bottom 10\% in our test set. The first column shows two input panoramas with their estimated layouts. The remaining columns show the ground-truth layout, our layouts, LED$^2$-Net's layouts, and LGT-Net's layouts in \textcolor{gtlayout}{blue}, \textcolor{ourlayout}{green}, \textcolor{ledlayout}{yellow}, \textcolor{lgtlayout}{red}, respectively. }
    \label{fig:comp_sota}
\end{figure}

In this experiment, we evaluate both the qualitative and quantitative performance of our model on the layout reconstruction task by comparing with baselines. The qualitative results are shown in \Figref{fig:comp_sota}. 
In short, our method produces more accurate layout reconstruction than LED$^2$-Net and LGT-Net.
Note that visual comparison with PSMNet is infeasible since the relevant code and data are not released. 
%In contrast with LED$^2$-Net and LGT-Net, our method could robustly predict layout reconstruction.
%
Our method also achieves the best performance against all the baselines across all evaluation metrics, as shown in \tabref{result_layout_comp}.
In the setting without ground-truth (GT) camera poses, our model shows an improvement over the PSMNet by $4.49\%$ for 2D IoU and 0.02 for $\delta^{i}$ without needing a (noisy) pose prior.
Please refer to the supplementary material for more visual comparisons.

% \jbox{\linewidth}{80pt}{Top layout visual results}
% \jbox{\linewidth}{80pt}{Average layout visual results}
% \jbox{\linewidth}{80pt}{Bad layout visual results}
% \jbox{0.3\linewidth}{80pt}{LED2Net result on our testing set}
% \jbox{0.3\linewidth}{80pt}{LGT-Net result on our testing set}
% \jbox{0.3\linewidth}{80pt}{Our result on our testing set}

\subsection{Evaluation on indoor panorama registration}
In this experiment, we compare {\networkName} with OpenMVG in terms of image registration quality on the indoor panoramas. As shown in~\tabref{result_reg_comp}, {\networkName} outperforms OpenMVG across all evaluation metrics.
The main problem of OpenMVG is that it has many failure cases where the algorithm returns no registration. Therefore, we show a separated comparison on the subset of panorama pairs where OpenMVG successfully returns a result. Note that even in these cases, our model beats OpenMVG with a noticeable improvement in angular errors.
Moreover, our model overwhelms OpenMVG when evaluating how many pairs are successfully registered within a certain error threshold.

\subsection{Ablation Study}
We conducted ablation studies to validate our method from different perspectives.

\heading{The influence of the number of cast rays.}
In this experiment, we start with the default setting of $N=256$ sample points along the $u$ coordinate and progressively add more samples during the inference without re-training the network. As shown in \tabref{result_abl_ray_layout} and \tabref{result_abl_ray_reg}, we obtain the best performance with the default setting on both the layout reconstruction and panorama registration. We conclude that re-training the network would be necessary for increasing the number of samples, but this will significantly increase computation time. 
%\james{we need an explanation why increasing the number of cast rays does not help...}

\begin{table}[!t]
    \caption{\textbf{Number of samples $N$ vs. registration accuracy.}}
    \label{tab:result_abl_ray_reg}
    \centering
    \resizebox{\textwidth}{!}{
        \begin{tabular}{lll|llll}
        \toprule
        & \multicolumn{2}{c}{SfM successful pair} & \multicolumn{4}{c}{full testing pair} \\
        \midrule
        $N$ & $\angularErr{t} \downarrow$ & $\angularErr{R} \downarrow$ &
        R-mAA$@5^{\circ} \uparrow$ & R-mAA$@10^{\circ} \uparrow$ & T-mAA$@5^{\circ} \uparrow$ & T-mAA$@10^{\circ} \uparrow$ \\ 
        \midrule
        256  & \textbf{6.9907} &\textbf{2.5514} &\textbf{0.8633} &\textbf{0.9378} &\textbf{0.7667} &\textbf{0.8854} \\ 
        512  & 16.9696 & 8.3937 & 0.6390 & 0.7183 & 0.5171 & 0.6416 \\ 
        1024  & 16.5873 & 7.2243 & 0.6825 & 0.7314 & 0.5901 & 0.6701 \\ 
        % 2048  & 17.7859 & 8.2320 & 0.6891 & 0.7309 & 0.6063 & 0.6787 \\
        % 4096  & - & - & - & - & - & - \\
        \bottomrule
        \end{tabular}
    }
\end{table}

\begin{table}[!t]
    \caption{\textbf{Number of samples $N$ vs. layout reconstruction accuracy.}}
    \label{tab:result_abl_ray_layout}
    \centering
    % \resizebox{\textwidth}{!}{
        \begin{tabular}{lllllll}
        \toprule
        & \multicolumn{3}{c}{w/ GT pose} & \multicolumn{3}{c}{w/o GT pose} \\
        \midrule
        $N$ & 2D IoU$\uparrow$ & $\delta^{i}$ $\uparrow$ & 3D IoU$\uparrow$ & 2D IoU$\uparrow$ & $\delta^i$ $\uparrow$ & 3D IoU$\uparrow$           \\ 
        \midrule
        256 & \tb{0.8449} &\tb{0.9603} & \tb{0.8211} &\tb{0.8026} &\tb{0.9452} & \tb{0.7816} \\ 
        512 & 0.7853 & 0.9472 & 0.7633 & 0.6917 & 0.8618 & 0.6730 \\ 
        1024 & 0.7619 & 0.9405 & 0.7407 & 0.6569 & 0.8437 & 0.6392 \\ 
        % 2048 & 0.7472 & 0.9366 & 0.7265 & 0.6309 & 0.8365 & 0.6138 \\ 
        % 4096 & - & - & - & - & - & - \\ 
        \bottomrule
        \end{tabular}
    % }
\end{table}

\heading{The effect of joint optimization architecture.}
In this experiment, we divided the layout prediction and the correspondence prediction into two individual models. Specifically, we use LED$^2$-Net as our layout prediction model and the vision transformer as our correspondence prediction model. As shown in \tabref{result_abl_joint_layout}, we obtain better accuracy using our joint prediction model on all the layout reconstruction metrics.

% \begin{table}[!t]
%     \caption{\textbf{Joint optimize architecture vs. layout reconstruction accuracy.}}
%     \label{tab:result_abl_joint_layout}
%     \centering
%     % \resizebox{\textwidth}{!}{
%         \begin{tabular}{lllllll}
%         \toprule
%         & \multicolumn{3}{c}{w/ GT pose} & \multicolumn{3}{c}{w/o GT pose} \\
%         \midrule
%         Method & 2D IoU$\uparrow$ & $\delta^{i}$ $\uparrow$ & 3D IoU$\uparrow$ & 2D IoU$\uparrow$ & $\delta^i$ $\uparrow$ & 3D IoU$\uparrow$ \\ 
%         \midrule
%         w/o joint optimization & 0.8364 & 0.9557 & 0.8131 & 0.7920 & 0.9413 & 0.7713 \\ 
%         w/ joint optimization & \tb{0.8449} &\tb{0.9603} & \tb{0.8211} &\tb{0.8026} &\tb{0.9452} & \tb{0.7816} \\
%         \bottomrule
%         \end{tabular}
%     % }
% \end{table}

\begin{table}[!t]
    \caption{\textbf{Joint optimize architecture vs. layout reconstruction accuracy.}}
    \label{tab:result_abl_joint_layout}
    \centering
    % \resizebox{\textwidth}{!}{
        \begin{tabular}{lllllll}
        \toprule
        & \multicolumn{3}{c}{w/ GT pose} & \multicolumn{3}{c}{w/o GT pose} \\
        \midrule
        Method & 2D IoU$\uparrow$ & $\delta^{i}$ $\uparrow$ & 3D IoU$\uparrow$ & 2D IoU$\uparrow$ & $\delta^i$ $\uparrow$ & 3D IoU$\uparrow$ \\ 
        \midrule
        w/o joint optimization & 0.8364 & 0.9557 & 0.8131 & 0.7920 & 0.9413 & 0.7713 \\ 
        w/ joint optimization & \tb{0.8449} &\tb{0.9603} & \tb{0.8211} &\tb{0.8026} &\tb{0.9452} & \tb{0.7816} \\
        \bottomrule
        \end{tabular}
    % }
\end{table}

\section{Conclusions}
\label{sec:conclusions}

We present a first complete solution for room layout reconstruction from a pair of panorama images. In contrast to previous work, i.e. PSMNet, we do not rely on an approximate registration but can register the two panorama images directly.
The major improvement over PSMNet comes from a novel Geometry-aware Panorama Registration Network (GPR-Net) that effectively tackles the wide baseline registration problem. We propose to exploit the layout geometry and compute fine-grained correspondences between the two layout boundaries, rather than directly computing the registration on global pixel-space.
The main limitation of our method is that the layout fusion block that processes two layouts is very simple. We recommend the development of learned fusion modules as major avenue for future work.

% \bibliography{iclr2023_conference}
% \bibliographystyle{iclr2023_conference}
\bibliography{ms}
\bibliographystyle{ms}

% \appendix
% \section{Appendix}

\end{document}